\documentclass[runningheads]{llncs}
\usepackage[T1]{fontenc}
\usepackage{graphicx}
\usepackage{booktabs}
\usepackage{tabularx}
\usepackage[table]{xcolor}
\definecolor{Gray}{gray}{0.9}
\usepackage{subcaption}
\usepackage{booktabs}
\usepackage{siunitx}
\usepackage{multirow}
\usepackage{amsmath}
\usepackage{amssymb}

\usepackage{float}

\begin{document}
\title{Active Reference Acquisition \\in Few-Shot Font Generation}
%
%


\author{Shinnosuke Matsuo\orcidID{0009-0007-2393-186X}\thanks{This work was conducted while the author was affiliated with Kyushu University.}} 
\authorrunning{S. Matsuo} 
\institute{NTT, Inc., Japan\\ \email{shinnosuke.matsuo@ntt.com}}
\maketitle
\begin{abstract}
Few-shot font generation aims to synthesize the remaining glyphs of a font given one or a few reference glyphs while preserving stylistic consistency, thereby supporting font designers in efficiently completing a typeface. Existing methods primarily focus on improving generation quality given a fixed reference set. However, when the current reference glyphs are insufficient to represent the target style, few-shot font generation may fail to produce satisfactory results. In practical scenarios, additional reference glyphs can often be obtained from the designer when necessary. Accordingly, we propose a new framework, \emph{Active Reference Acquisition in Few-Shot Font Generation}, in which the model sequentially decides which character to acquire next as an additional reference. Furthermore, we propose a reference part-coverage-based acquisition function to efficiently query the designer. Motivated by the observation that font styles are well characterized by local structural parts, we represent each glyph using a histogram of local features and select query characters that maximize the expected part coverage of the reference set. By prioritizing characters that contain parts not yet covered by the current references, the proposed method progressively expands the diversity of visual parts in the reference set. As a result, generation quality is improved with fewer queries. Experiments on the Google Fonts dataset demonstrate that the proposed method achieves higher generation quality than random querying and reference-agnostic baselines. The code is available at https://github.com/matsuo-shinnosuke/ActiveRef-FontGen.
\keywords{Few-shot font generation  \and Active reference acquisition.}
\end{abstract}
\section{Introduction}

\begin{figure}[t]
    \centering
    \includegraphics[width=\linewidth]{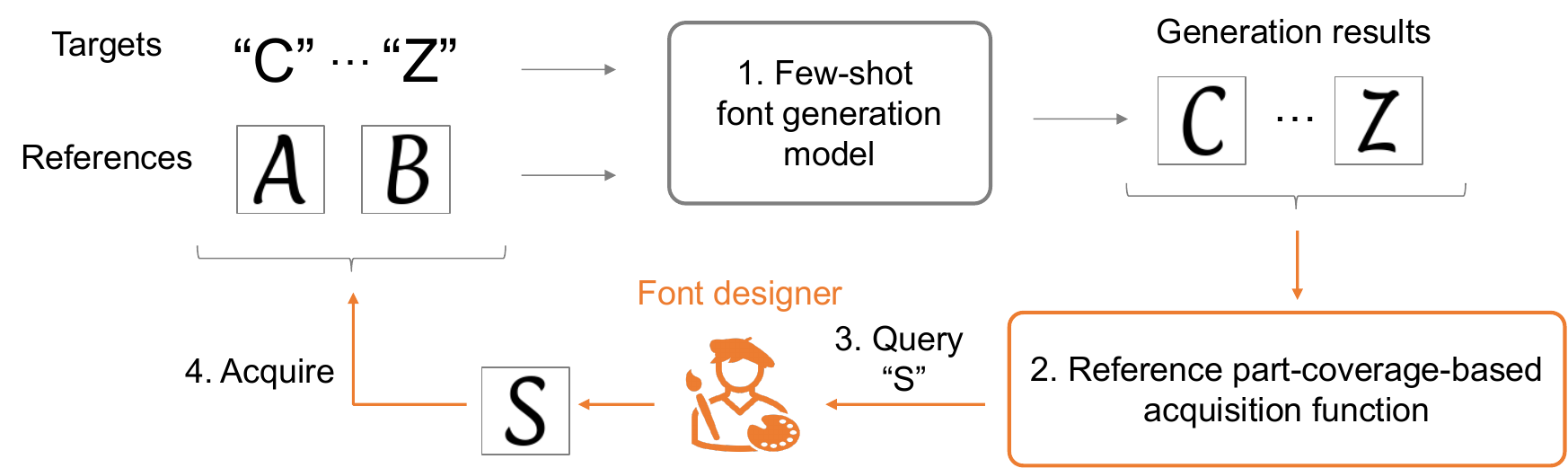}
    \caption{Active reference acquisition in few-shot font generation.}
    \label{fig:problem_setting}
\end{figure}

Designing a font is a highly time-consuming and labor-intensive process that relies heavily on expert craftsmanship.
Even for the Latin alphabet, a designer must create at least 26 uppercase glyphs, and in practice many more, including lowercase letters, numerals, punctuation, and symbols.
Moreover, all glyphs within a typeface must share consistent fine-grained stylistic characteristics, such as stroke terminals, contrast, curvature, and proportions.
Ensuring such consistency across all characters imposes a significant burden on designers.

To alleviate this burden, the problem of \emph{few-shot font generation} has been studied~\cite{azadi2018multicontent,cha2020dmfont,park2021lffont,park2021mxfont,he2024difffont}.
This task aims to synthesize a complete alphabet from one or a few reference glyphs created by a designer. 
For example, as illustrated in the upper part of Fig.~\ref{fig:problem_setting}, given a small set of reference characters such as ``A''--``C'', the model generates the remaining characters ``D''--``Z'' while preserving stylistic consistency.

Recent few-shot font generation methods are typically realized using either GAN-based or diffusion-based generative models. 
GAN-based approaches focus on disentangling content (character identity) and style (font appearance), and several works further leverage compositional structures or localized style representations to capture fine-grained details~\cite{azadi2018multicontent,cha2020dmfont,park2021lffont,park2021mxfont,tang2022finegrained}. 
More recently, diffusion-based approaches have demonstrated strong performance. 
In particular, Diff-Font~\cite{he2024difffont} provides a robust framework for one-shot font synthesis, and subsequent work has extended this line of research~\cite{yang2024fontdiffuser}. 
Despite architectural differences, these methods typically assume a fixed reference glyph set and focus on improving the accuracy of generating non-reference glyphs conditioned on the given references.

However, in practical design workflows, there are situations where the current reference glyph images are insufficient and few-shot font generation fails to produce satisfactory results. In such cases, it is often possible to query the designer for additional reference glyphs as needed. This is particularly important for complex or decorative fonts, where a single or a small number of reference glyph images cannot fully represent the entire style. Although adding more references can potentially improve generation quality, the designer’s time is limited; therefore, it is necessary to carefully select which glyph to query next.

In this paper, we propose \emph{Active Reference Acquisition} in few-shot font generation, a designer-in-the-loop formulation that enables efficient few-shot font generation with a limited number of queries. An overview is shown in Fig.~\ref{fig:problem_setting}. 
First, given one or a few initial reference glyphs for a target font, an existing few-shot font generation model is used to synthesize the remaining target characters. Based on the generated results, an acquisition function is computed to estimate the value of querying each candidate glyph from the designer. The glyph with the highest acquisition value is then queried and added to the reference set. By iteratively repeating this process, generation performance can be improved more efficiently than with naive policies such as random querying.

\begin{figure}[t]
    \centering
    \begin{subfigure}{\textwidth}
        \centering
        \includegraphics[width=0.85\linewidth]{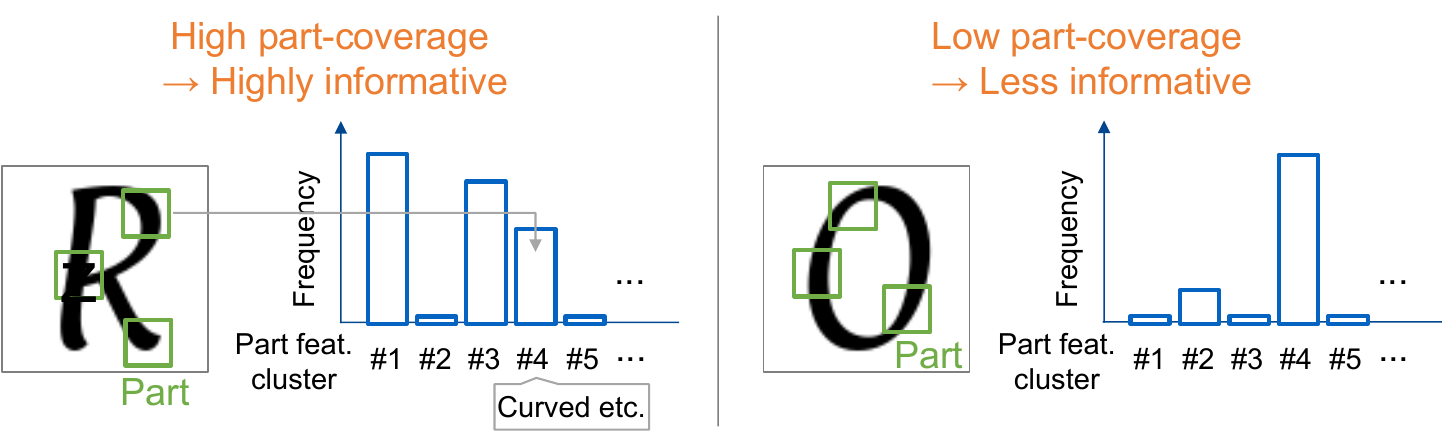}
        \vspace{-2mm}
        \caption{Part-coverage representation}
    \end{subfigure}
    
    \begin{subfigure}{\textwidth}
        \centering
        \vspace{3mm}
        \includegraphics[width=0.9\linewidth]{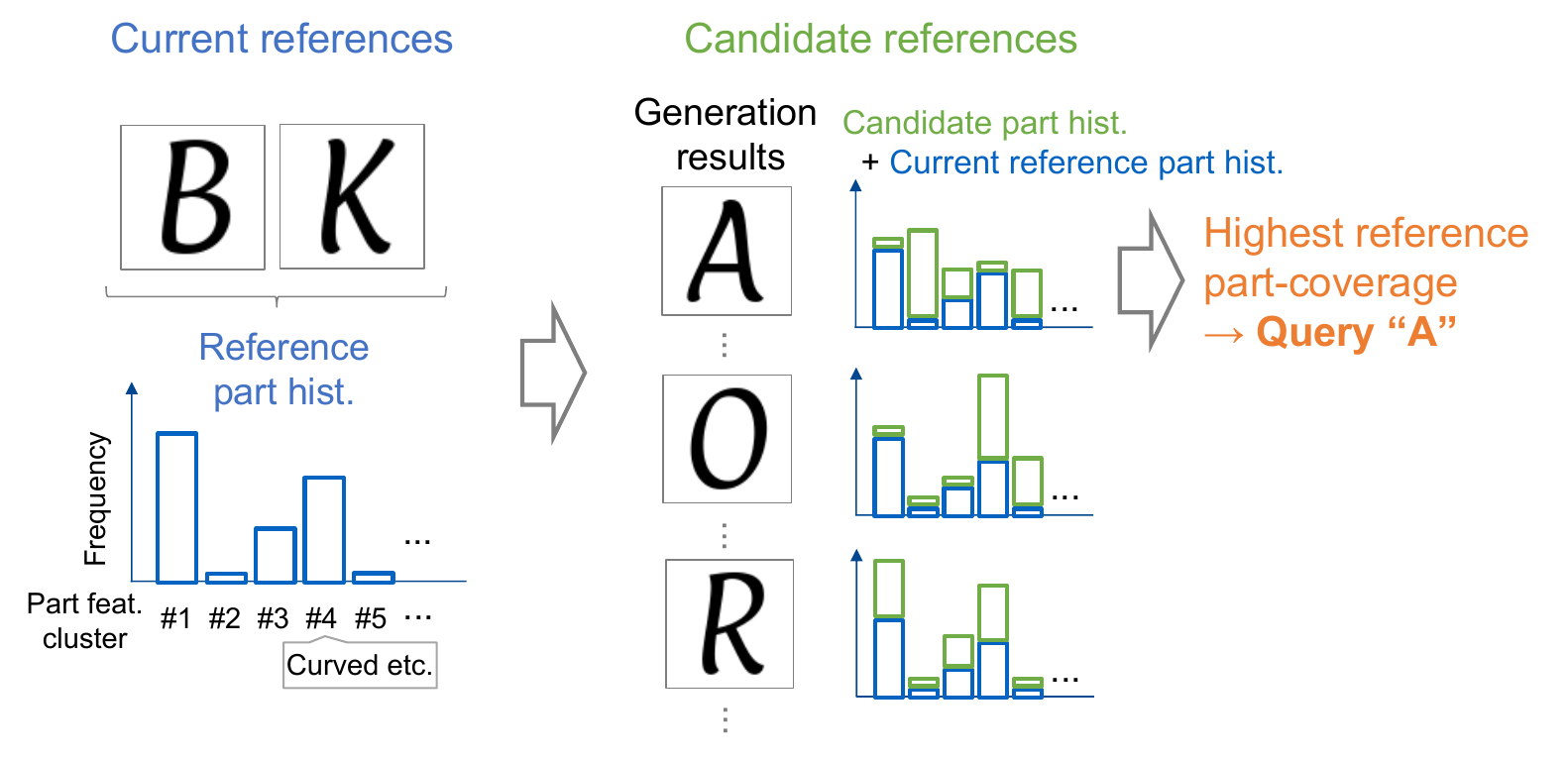}
        \vspace{-2mm}
        \caption{Query selection that considers both candidate and reference part coverage.}
    \end{subfigure}
        \vspace{-5mm}
    \caption{Overview of the reference part-coverage-based acquisition function.}
    \label{fig:acquisition}
\end{figure}

To address this problem, we propose an acquisition function that focuses on local structural parts of glyphs. 
Font styles are often characterized by distinctive parts such as vertical strokes, diagonal strokes, curves, and stroke terminals. 
Therefore, in few-shot font generation, it is reasonable to assume that reference glyphs covering a diverse set of parts are more informative.

Accordingly, as illustrated in Fig.~\ref{fig:acquisition}(a), we define the part coverage of each character. 
Specifically, local features corresponding to glyph parts are extracted using descriptors such as SIFT~\cite{lowe2004distinctive}, followed by clustering to construct a histogram of clustered part features.
The part coverage is defined as the entropy of the normalized histogram. 
A higher value indicates that the character contains a diverse set of parts, such as the vertical strokes, curves, diagonals, and terminals found in ``R'', whereas a lower value indicates limited diversity, as in characters dominated mainly by curves.

The most naive strategy is to query characters solely based on their individual part coverage. 
However, this approach suffers from an important limitation. 
For example, when ``B'' and ``K'' are already included in the reference set, a character such as ``R'' may have high part coverage on its own; nevertheless, many of its parts are already present in ``B'' and ``K'', making it less valuable to add as an additional reference.

To address this issue, as illustrated in Fig.~\ref{fig:acquisition}(b), we propose a \emph{reference part-coverage-based acquisition function} that also takes into account the part histogram of the current reference set. 
Specifically, for each unacquired candidate character, we compute its part histogram from the generated glyph image, add it to the current reference part histogram, and evaluate the expected reference part coverage if that character were acquired. 
This expected reference part coverage is used as the acquisition value.
By doing so, the method efficiently queries characters that contain parts not yet covered by the current reference set, progressively expanding the diversity of reference parts and thereby improving the accuracy of few-shot font generation. 

Experiments on Google Fonts demonstrate that the proposed method achieves higher generation quality with fewer queries compared to random querying, fixed query orders derived from training data, and query strategies that ignore the current reference set.

Our main contributions are summarized as follows.
\begin{itemize}
    \item We introduce \emph{Active Reference Acquisition} in few-shot font generation, a new framework that enables actively querying a designer for additional reference glyphs.
    \item We propose a \emph{reference part-coverage-based acquisition function} that selects query characters by maximizing the expected part coverage of the reference set, explicitly accounting for redundancy with existing references.
    \item We demonstrate through experiments on the Google Fonts dataset that the proposed method achieves higher generation quality with fewer queries than random querying and reference-agnostic baselines.
\end{itemize}

\section{Related Work}
\label{sec:related_work}

\subsection{Few-Shot Font Generation}
Few-shot font generation aims to synthesize a complete font set (e.g., all glyphs in a character set) from one or a few reference glyph images designed by an expert. A common formulation treats the standard glyphs in a source font as \emph{content} inputs and transfers the \emph{style} extracted from the reference glyphs to the remaining characters, seeking both high-fidelity stylization and consistency across the generated alphabet.

Early and widely studied approaches build on conditional image-to-image translation and GAN-based frameworks. Multi-Content GAN~\cite{azadi2018multicontent} addresses ornamented fonts and enables synthesis of unobserved glyphs from limited references. Subsequent works improve few-shot generalization by exploiting compositional structures and localized style cues: DM-Font~\cite{cha2020dmfont} introduces a dual-memory mechanism to leverage script compositionality, while LF-Font~\cite{park2021lffont} and MX-Font~\cite{park2021mxfont} represent style in a localized/component-wise manner to better capture fine details. Tang \textit{et al.}~\cite{tang2022finegrained} further emphasize fine-grained local styles and propose cross-attention to align local style patterns between reference and target glyphs.

More recently, diffusion-based models have also been explored for one-/few-shot font generation. Diff-Font~\cite{he2024difffont} formulates one-shot font synthesis with diffusion to improve training stability and fidelity, and FontDiffuser~\cite{yang2024fontdiffuser} proposes diffusion-based denoising with multi-scale content aggregation and style contrastive learning for robust imitation. 

Importantly, despite architectural differences, the standard few-shot font generation setting largely assumes that the reference glyph set is fixed and does not consider updating it, focusing instead on maximizing generation quality given those references.
In contrast, our work focuses on \emph{which additional references to query a designer for} in order to improve generation efficiency, and can be combined with existing GAN- or diffusion-based generators.

\subsection{Active Learning}
Active learning (AL) studies how to reduce annotation cost by \emph{selectively} querying a human annotator for labels of informative examples. In the standard pool-based setting, an algorithm iteratively chooses unlabeled samples according to an acquisition strategy (e.g., uncertainty or expected utility) and obtains their labels to improve a model with fewer labeled instances than random sampling~\cite{settles2009survey}.

Classic AL strategies include uncertainty sampling and its variants for text and classification problems~\cite{lewis1994sequential,lewis1994heterogeneous}, as well as committee-based methods such as query-by-committee~\cite{seung1992qbc}. Beyond heuristics, Bayesian and decision-theoretic views lead to criteria that minimize expected posterior uncertainty or expected error under statistical models~\cite{cohn1996active}. In deep learning, AL has been combined with Bayesian approximations and diversity-aware batch selection, such as deep Bayesian active learning~\cite{gal2017dbal}, core-set selection for CNNs~\cite{sener2018coreset}, and batch selection via gradient embeddings (BADGE)~\cite{ash2020badge}.

Our setting differs from conventional AL in two key ways. First, AL primarily targets \emph{training-time} data efficiency, querying labels to update a predictive model. Second, in conventional AL, queries correspond to requesting \emph{labels} for instances drawn from an unlabeled data pool. In contrast, we query a designer \emph{at inference time} for additional \emph{reference glyph images} to better condition a font generator, with the goal of improving generation quality under a limited number of queries. Thus, while both AL and our approach use acquisition functions, the queried object (labels vs.\ reference images), the timing (training vs.\ inference), and the objective (classification accuracy vs.\ generative fidelity/consistency) are fundamentally different.

\subsection{Active Feature Acquisition}
Active feature acquisition (AFA) studies decision-making under incomplete observations where acquiring additional feature values incurs a cost. Given a partially observed instance, an agent sequentially decides which feature to acquire next, or when to stop acquisition and make a prediction, aiming to maximize predictive performance with a limited number of acquisitions.

Recent AFA methods develop principled acquisition criteria and learn acquisition policies directly from data. AFASMC~\cite{huang2018afasmc} combines active querying with supervised matrix completion to exploit feature correlations, while RL-based approaches jointly learn when to acquire features and when to predict~\cite{shim2018jafa}. Generative modeling has also been used to estimate information gain and guide acquisitions, including Partial-VAE-based information acquisition (EDDI)~\cite{ma2019eddi} and generative surrogate models for policy learning~\cite{li2021gsm}.

Although AFA is conceptually related to our approach in that both aim to maximize downstream performance with a limited number of queries, most AFA work focuses on \emph{tabular} feature values and \emph{classification/regression} objectives for a single instance. In contrast, we study a \emph{generative} setting where the acquired information is a new reference glyph image for a font style, and the benefit is measured by improved generation quality across multiple target characters.

\section{Problem Setting: Active Reference Acquisition}
\label{sec:problem_setting}

In this section, we formalize \emph{Active Reference Acquisition} for few-shot font generation, a new problem setting in which the system is allowed to actively acquire additional reference glyphs from a designer.

Consider a target character set of size $N$ (e.g., $N=26$ for uppercase Latin letters A--Z).
Let $\mathcal{V}=\{1,\ldots,N\}$ denote the set of character identities.
For a given font, let $x_c$ denote the ground-truth glyph image of character $c \in \mathcal{V}$.

In few-shot font generation, the system is initially provided with a small set of reference character identities 
$\mathcal{S}_0 \subset \mathcal{V}$, where $|\mathcal{S}_0| = K_0$ and $K_0$ is typically one or a few.
Given the corresponding reference glyph images $\{x_c \mid c \in \mathcal{S}_0\}$, 
a generator synthesizes glyph images for the remaining characters $\mathcal{V} \setminus \mathcal{S}_0$.

In practical design workflows, it is often possible to request additional reference glyphs when the current references are insufficient.
We therefore consider an interactive setting in which the system iteratively selects a new character identity to query.
At iteration $t$, let $\mathcal{S}_t \subset \mathcal{V}$ denote the current set of reference character identities.
The system selects a query character $q_t \in \mathcal{V} \setminus \mathcal{S}_t$, obtains its glyph image $x_{q_t}$ from the designer, and updates the reference set by setting $\mathcal{S}_{t+1} = \mathcal{S}_t \cup \{q_t\}$.
This process continues under a query budget, resulting in a final reference set $\mathcal{S}_T$ of size $K$.

The objective is to design a querying policy that selects a sequence of characters so as to maximize the final generation quality on the remaining characters $\mathcal{V} \setminus \mathcal{S}_T$.
Generation quality is measured by a task-specific metric (e.g., SSIM or LPIPS) between generated glyph images and the corresponding ground-truth glyph images.

\section{Proposed Method: Reference Part-Coverage-based Acquisition Function}
\label{sec:method}

In this section, we propose an efficient strategy for \emph{Active Reference Acquisition} in few-shot font generation.
A font style is characterized not only by the global appearance of a typeface, but also by the consistency of fine-grained local details, such as stroke terminals, junctions, curves, and decorative parts.
Therefore, in few-shot font generation, it is important that the reference set covers a wide range of parts appearing across characters within the font.
Based on this idea, we design an acquisition function that selects the next query character using the \emph{predicted} part coverage of the reference set after adding that character.

We first describe how to represent part coverage for a glyph image in Sec.~\ref{subsec:part_coverage_repr}.
Then, in Sec.~\ref{subsec:rpc_acq}, we introduce our \emph{Reference part-coverage-based Acquisition Function}, which selects the character identity whose acquisition is expected to maximize reference part coverage.

\subsection{Part-Coverage Representation}
\label{subsec:part_coverage_repr}

We describe how to represent the part coverage of each glyph image.
Since fine-grained local details are crucial for few-shot font generation, characters that cover diverse parts can be particularly informative as references.
For example, as illustrated in Fig.~\ref{fig:acquisition}(a), some characters (e.g., \emph{R}) contain multiple parts such as vertical strokes, curves, diagonals, and decorative elements (e.g., distinctive terminals), whereas other characters (e.g., \emph{O}) are dominated by a smaller variety of parts (mainly curves).

Our representation consists of three steps.
First, given a glyph image $x$, we extract local descriptors from the image (e.g., SIFT descriptors, or deep local descriptors from an intermediate feature map of a trained style encoder).
Second, at iteration $t$, we collect local descriptors from the available glyph images (the current references and the model-generated glyphs for the remaining characters) and run $k$-means to obtain $M$ clusters. We then compute $h(x)$ by assigning descriptors to these clusters.
Intuitively, each cluster is expected to correspond to a recurring part pattern, such as straight strokes, curved strokes, junctions, or decorative textures.
Third, we represent each glyph image $x$ by a histogram $h(x)\in\mathbb{R}^{M}_{\ge 0}$, where the $m$-th bin counts how many local descriptors are assigned to cluster $m$.
We use this histogram as the part-coverage representation.

To measure part coverage as a scalar, we compute the entropy of the normalized histogram.
Concretely, we normalize $h(x)$ by dividing each bin by the total number of descriptors (i.e., by $\sum_{m=1}^{M} h_m(x)$), and compute Shannon entropy.
For convenience, we define the entropy of a (nonnegative) histogram vector $h$ as
\begin{equation}
\label{eq:entropy_hist}
H(h)\;=\;-\sum_{m=1}^{M} p_m \log p_m,
\qquad
p_m \;=\; \frac{h_m}{\sum_{m'=1}^{M} h_{m'}}.
\end{equation}
If a glyph is dominated by a small subset of parts, the normalized histogram is peaked and $H(h(x))$ becomes small.
In contrast, if a glyph contains a wider variety of parts, the histogram becomes more spread out and $H(h(x))$ becomes larger.
We use $H(h(x))$ as a proxy for how well the glyph covers diverse parts.

\subsection{Reference Part-Coverage-based Acquisition Function}
\label{subsec:rpc_acq}

Using the part-coverage representation, a naive strategy would be to query the character whose individual part coverage $H(h(\cdot))$ is the largest.
However, this can be suboptimal because a candidate glyph may be redundant when its parts are already covered by the current references.
For example (Fig.~\ref{fig:acquisition}), suppose that the current reference set already contains the characters ``B'' and ``K''.
Although the character ``R'' exhibits high individual part coverage due to its multiple parts, many of these parts (e.g., vertical strokes and curved components) are already well represented by ``B'' and ``K'', and thus adding ``R'' provides limited additional coverage.
In contrast, a character such as ``A'', which contains distinctive diagonal strokes and a sharp apex at the top, introduces parts that are not sufficiently covered by the existing references.
Even if ``A'' has lower individual part coverage than ``R'', its addition increases the diversity of parts in the reference set and is therefore expected to improve few-shot font generation more effectively.

To address this, we propose a \emph{Reference part-coverage-based acquisition function} that explicitly accounts for the current reference set.
The goal of this acquisition function is to efficiently query characters that contain parts not included in the current reference set, thereby sequentially expanding the variety of parts represented by the references.
Furthermore, when multiple candidate characters contain parts not present in the reference set, the function is designed to select the one with higher individual part coverage.
In this way, the acquisition simultaneously considers (i) whether a candidate character includes parts absent from the current references and (ii) how rich the candidate character is in terms of part coverage.

At iteration $t$, let $\mathcal{S}_t \subset \mathcal{V}$ be the current set of reference character identities.
We aggregate the part histograms of the current references as
\begin{equation}
h_{\mathrm{ref}}^{(t)} \;=\; \sum_{c\in\mathcal{S}_t} h(x_c).
\end{equation}
For each unqueried character $c \in \mathcal{V}\setminus\mathcal{S}_t$, the ground-truth glyph image cannot be accessed before acquisition.
Therefore, we approximate its part coverage using the glyph image synthesized by the current few-shot generator.
We denote this generated glyph by $\hat{x}_c^{(t)}$ and compute its part histogram $h(\hat{x}_c^{(t)})$.

We then select the next query character by maximizing the entropy (i.e., part coverage) of the \emph{predicted} reference histogram after adding $c$:
\begin{equation}
\label{eq:acq_entropy}
q_t \;=\; \arg\max_{c\in\mathcal{V}\setminus\mathcal{S}_t}
H\!\left(h_{\mathrm{ref}}^{(t)} + h(\hat{x}_c^{(t)})\right).
\end{equation}
After querying $q_t$ and obtaining its glyph image from the designer, the reference set is updated and the procedure repeats until reaching the maximum number of allowed queries.

\section{Experiments}

\subsection{Dataset}
\label{subsec:dataset}

To evaluate the effectiveness of our approach, we use the Google Fonts dataset.\footnote{https://github.com/google/fonts}
In all experiments, we focus on the 26 uppercase Latin letters (A--Z).
From the Google Fonts collection, we extracted all fonts that contain the complete set of uppercase letters A--Z, resulting in 3,759 fonts in total.
Each font is associated with a family label, and the extracted set contains 1,978 distinct families.
Each glyph is rendered as a grayscale image of size $80 \times 80$ pixels.
To minimize similarity between training and test fonts, we split the dataset by family rather than by individual fonts, since fonts within the same family often share highly similar design characteristics.
Specifically, we partition the families into training and test sets at the family level, assigning 1,582 families (2,867 fonts) to the training set and the remaining 396 families (892 fonts) to the test set.

\subsection{Implementation Details}
\label{subsec:impl}

\subsubsection{Few-Shot Font Generation Model}
\label{subsubsec:generator}

As the base few-shot font generation model, we adopt Diff-Font~\cite{he2024difffont}, a conditional diffusion-based approach for one-shot font generation.
Diff-Font conditions the diffusion model on the target character identity and a style representation extracted from reference glyphs.
The original Diff-Font framework targets Chinese and Korean font generation and incorporates explicit component information as an additional condition.

In our setting, since we focus on uppercase Latin letters, we condition the generator only on the target character identity and the extracted style feature, and we omit component-related conditions.
Since Diff-Font is originally designed for one-shot font generation, we extend it to the few-shot setting.
When multiple reference glyphs are available, we extract a style feature from each reference glyph independently and aggregate them by mean pooling to obtain a single style representation.
This aggregated style feature is then used as the conditioning input to the diffusion model.

We train the generator on the training split of the Google Fonts dataset using AdamW with a learning rate of $1 \times 10^{-4}$ for 40{,}000 optimization steps.
During training, we sample $K \in \{1,2,3,4\}$ reference glyphs and aggregate their style features by mean pooling.

\subsubsection{Style Encoder}
\label{subsubsec:style_encoder}

We train a CNN-based style encoder to extract style features from reference glyph images.
The encoder is trained via a font classification task: for each font, we assign the same font label to its glyph images (A--Z) and optimize a standard multi-class classification objective.
We split 20\% of the training data as a validation set.
We train the encoder using AdamW with learning rate $3 \times 10^{-4}$ for 400 epochs, and select the checkpoint that achieves the best validation accuracy.

\subsubsection{Experimental Protocol}
\label{subsubsec:exp_protocol}

We follow a standard few-shot evaluation protocol.
For each test font, we start from a single reference glyph, i.e., $|\mathcal{S}_0|=1$.
Following the setup used in Diff-Font~\cite{he2024difffont}, the initial reference character is selected uniformly at random from the 26 uppercase letters.
To ensure fair comparisons, we use the same initial reference character for all methods for each font.
Starting from the initial reference, we iteratively acquire additional reference characters up to eight references in total.
We report generation performance at $K\in\{1,2,4,8\}$ shots.

We use SIFT descriptors~\cite{lowe2004distinctive} as local part features and cluster them into $M=32$ groups using the $k$-means algorithm to construct the part histogram representation used by the acquisition function.

\subsection{Evaluation Metrics}
\label{subsec:metrics}

We evaluate generation quality using SSIM~\cite{wang2004ssim}, RMSE, and LPIPS~\cite{zhang2018lpips}.
SSIM measures structural similarity between the generated glyph and the ground-truth glyph, with higher values indicating better structural consistency.
RMSE measures the pixel-wise reconstruction error, where lower values indicate more accurate reproduction.
LPIPS measures perceptual similarity based on deep features, where lower values indicate that the generated glyph is perceptually closer to the ground truth.

All metrics are computed between generated glyph images and the corresponding ground-truth glyph images and are averaged over the target characters (i.e., characters not included in the reference set) and over all test fonts.


\begin{table}[t]
\centering
\caption{Top-8 query characters used by the \emph{Fixed} baseline. The global acquisition order is derived from the training set by ranking characters according to their part coverage (entropy of the normalized part histogram) for each font and then averaging the ranks across fonts.}
\label{tab:fixed_top8}

\setlength{\tabcolsep}{7pt}
\begin{tabular}{
@{\hspace{1em}}
l c c c c c c c c
@{\hspace{1em}}
}
\toprule
\textbf{Rank} & \textbf{1} & \textbf{2} & \textbf{3} & \textbf{4} & \textbf{5} & \textbf{6} & \textbf{7} & \textbf{8} \\
\midrule
Query character 
& ``R'' & ``A'' & ``E'' & ``F'' & ``K'' & ``B'' & ``H'' & ``P'' \\
Average rank 
& 4.62 & 4.95 & 6.19 & 7.65 & 8.20 & 8.97 & 9.18 & 9.89 \\
Average part coverage 
& 2.48 & 2.46 & 2.36 & 2.27 & 2.24 & 2.25 & 2.21 & 2.19 \\
\bottomrule
\end{tabular}
\end{table}

\subsection{Baselines and Ablation}
\label{subsec:baselines}

To demonstrate the effectiveness of our method, we compare against two baselines and one ablation under the same number of query steps.
    \par\noindent \textbf{Random:}
    At each iteration $t$, the next query character is selected uniformly at random from the unqueried set $\mathcal{V}\setminus\mathcal{S}_t$.

    \par\noindent\textbf{Fixed:}
    A global acquisition order is precomputed from the training set and then applied uniformly to all test fonts; that is, the same sequence of characters is queried for every test font regardless of its style.
    Concretely, for each training font, we compute the part coverage (entropy of the normalized part histogram) for each character and obtain a per-font ranking.
    These rankings are then averaged across all training fonts to derive a single global order.
    Table~\ref{tab:fixed_top8} shows the top-8 characters in this order.
    During evaluation, characters are acquired following this fixed order; if a character is already included in the current reference set (e.g., as the initial reference), it is skipped and the next character in the sequence is selected.
    
    \par\noindent\textbf{Ours w/o Reference Part Coverage (RPC): }
    This ablation evaluates the importance of accounting for the current reference set in our acquisition function.
    Instead of maximizing the reference part coverage, this variant queries the character with the largest individual part coverage among unqueried characters.
    Specifically, at iteration $t$, it selects
    \begin{equation}
    \label{eq:acq_wo_rpc}
    q_t \;=\; \arg\max_{c\in\mathcal{V}\setminus\mathcal{S}_t} H\!\left(h(\hat{x}_c^{(t)})\right).
    \end{equation}
    Unlike the Fixed strategy, which follows a static global order shared across all fonts, this variant dynamically selects characters depending on each test font. However, it ignores redundancy between the candidate character and the existing references, and therefore serves as a naive baseline for our reference-aware acquisition.



\begin{figure}[t]
\vspace{3mm}
    \centering
    \begin{subfigure}{0.32\textwidth}
        \centering
        \includegraphics[width=\linewidth]{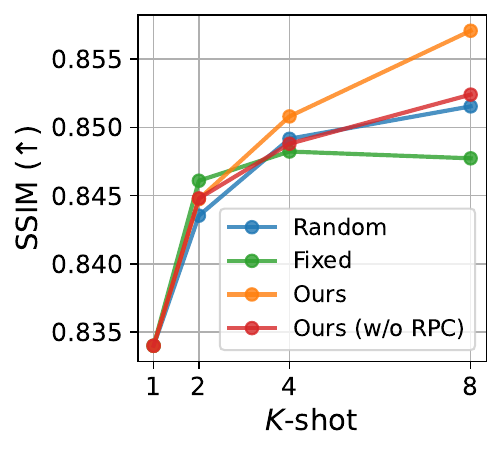}
        \vspace{-6mm}
        \caption{SSIM}
    \end{subfigure}
    \begin{subfigure}{0.32\textwidth}
        \centering
        \includegraphics[width=\linewidth]{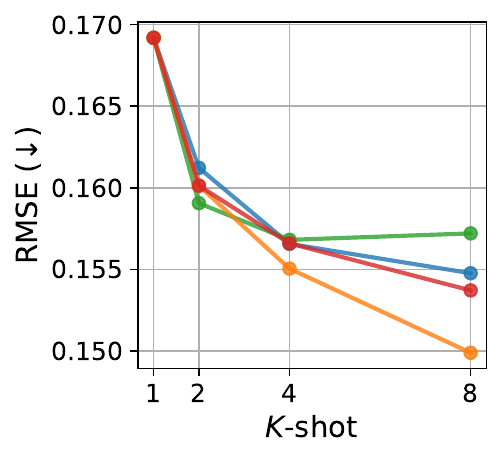}
        \vspace{-6mm}
        \caption{RMSE}
    \end{subfigure}
    \begin{subfigure}{0.32\textwidth}
        \centering
        \includegraphics[width=\linewidth]{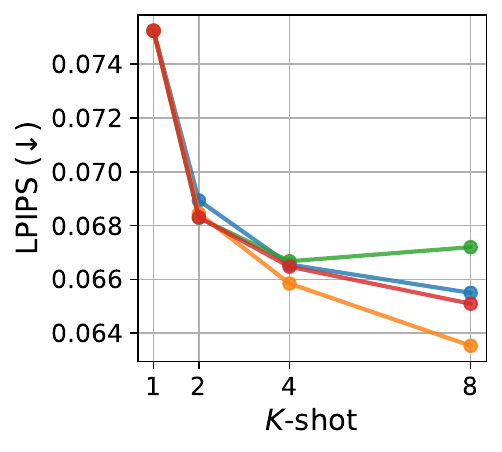}
        \vspace{-6mm}
        \caption{LPIPS}
    \end{subfigure}
        \vspace{-2mm}
\caption{Comparison with baselines and an ablation on Google Fonts.
}
\label{fig:comparison}
\end{figure}

\subsection{Comparative Results}
\label{subsec:results}

Fig.~\ref{fig:comparison} summarizes the quantitative comparison with two baselines and one ablation.
All methods start from the same single reference ($K=1$) and sequentially acquire additional reference characters.
We evaluate few-shot font generation at $K\in\{1,2,4,8\}$ using SSIM, RMSE, and LPIPS.

Our method consistently outperforms all three competing methods at $K=4$ and $K=8$ across all evaluation metrics.
Comparing with \emph{Random} shows that actively querying references is more effective than selecting characters at random.

Compared to \emph{Fixed}, which follows a static order derived from training fonts, our method dynamically selects query characters based on the target font and the current reference set, leading to superior performance.

Finally, compared with the ablation \emph{Ours w/o Reference Part Coverage (RPC)}, the results indicate that selecting characters solely based on their individual part coverage is insufficient.
Explicitly accounting for both the candidate's part coverage and the existing coverage of the reference set is essential for effective acquisition.

The difference at $K=2$ is relatively small.
With only one initial reference, the reference set contains very limited parts, and many candidate characters can introduce new parts.
As a result, adding one additional reference improves performance regardless of the specific character selected.
In contrast, once the reference set grows (e.g., $K=4$ or $K=8$), more parts are already covered, and effective querying requires identifying what remains missing.
This explains why our reference part-coverage-based acquisition function becomes more advantageous as $K$ increases.

\begin{figure}[t]
    \centering
    \includegraphics[width=\linewidth]{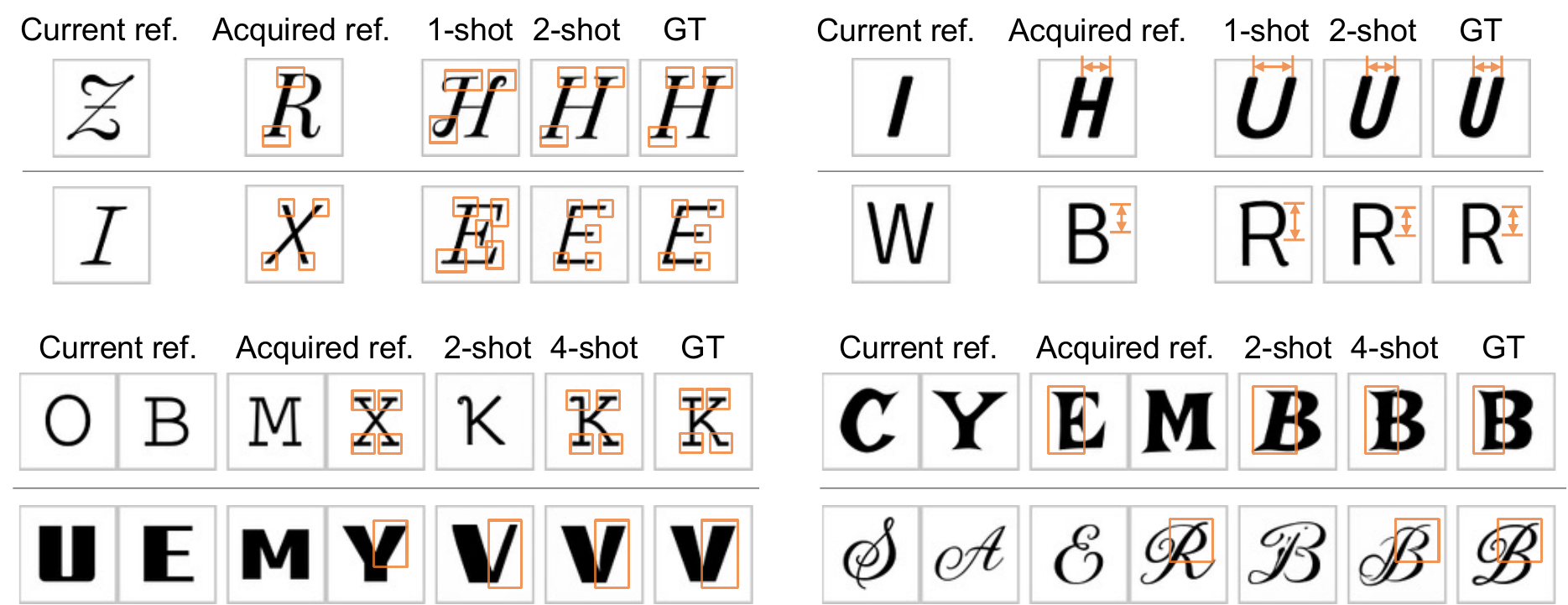}
\caption{Qualitative examples of generation improvements by our method.
}
\label{fig:generated_images}
\end{figure}

\subsection{Qualitative Results}
\label{subsec:qualitative}

Fig.~\ref{fig:generated_images} shows qualitative examples where the proposed method improves generation quality after acquiring additional reference glyphs.
For each example, we visualize the current reference glyph(s), the acquired reference glyph(s) selected by our acquisition function, the generated glyph before and after acquisition, and the ground-truth (GT).
We highlight regions of interest using orange boxes.

In the top row (1-shot $\rightarrow$ 2-shot), we observe improvements in fine-grained details such as terminals and stroke shapes.
For instance, when the current reference is ``Z'', acquiring ``R'' improves the vertical stroke and terminal details of the generated ``H'', which are difficult to infer from ``Z'' alone.
Similarly, acquiring ``X'' refines the terminal/corner details of ``E'' compared to the 1-shot result.
The two examples on the right further demonstrate improvements in inter-stroke spacing: acquiring ``H'' helps correct the spacing of ``U'', and acquiring ``B'' improves the spacing and curved structure of ``R''.

The bottom row shows 2-shot $\rightarrow$ 4-shot cases, where two additional references are acquired.
Similarly, the newly acquired references introduce parts not present in the current reference set, such as distinctive stroke terminals, vertical strokes, and diagonal strokes.
As a result, the generated glyphs for ``K'', ``V'', and ``B'' become closer to GT.
Overall, these qualitative results support our motivation that actively acquiring complementary references that contain previously unseen parts leads to improved few-shot font generation.

\begin{figure}[t]
    \begin{subfigure}{0.32\textwidth}
        \centering
        \includegraphics[width=0.9\linewidth]{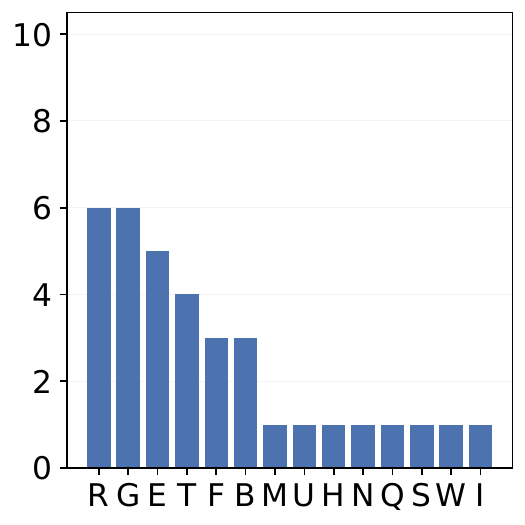}
        \vspace{-2mm}
        \caption{Reference: ``A''}
    \end{subfigure} 
    \begin{subfigure}{0.32\textwidth}
        \centering
        \includegraphics[width=0.9\linewidth]{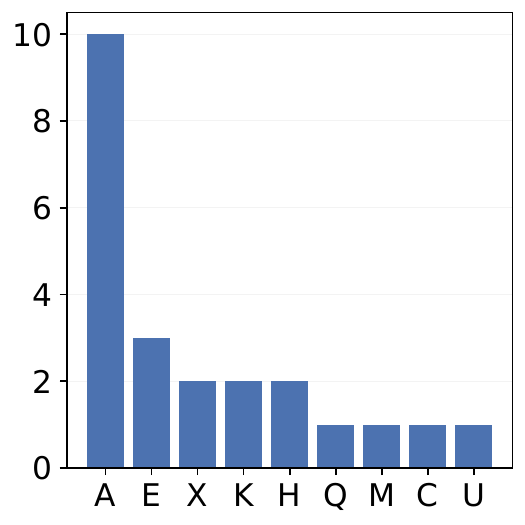}
        \vspace{-2mm}
        \caption{Reference: ``B''}
    \end{subfigure}
    \begin{subfigure}{0.32\textwidth}
        \centering
        \includegraphics[width=0.9\linewidth]{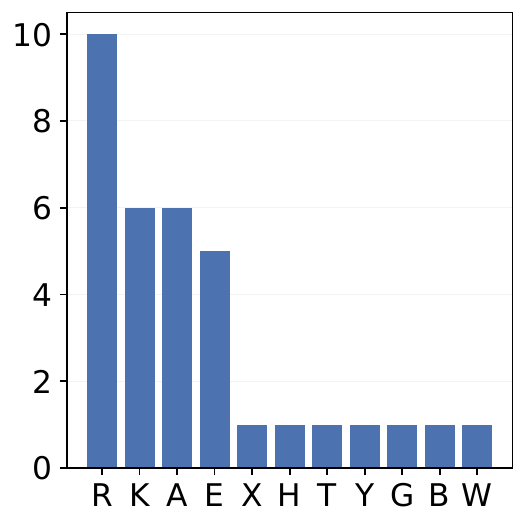}
        \vspace{-2mm}
        \caption{Reference: ``C''}
    \end{subfigure}
    \vspace{-2mm}
    \caption{Histogram of the additionally acquired reference character aggregated over all test fonts for different single-reference settings ($K = 1$).}
    \label{fig:acquired_oneshot}
\end{figure}

\begin{figure}[t]
    \begin{subfigure}{0.32\textwidth}
        \centering
        \includegraphics[width=0.9\linewidth]{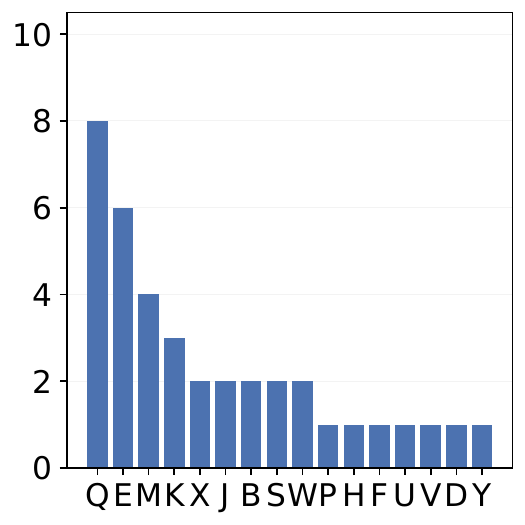}
        \vspace{-2mm}
        \caption{Reference: ``A'', ``R''}
    \end{subfigure} 
    \begin{subfigure}{0.32\textwidth}
        \centering
        \includegraphics[width=0.9\linewidth]{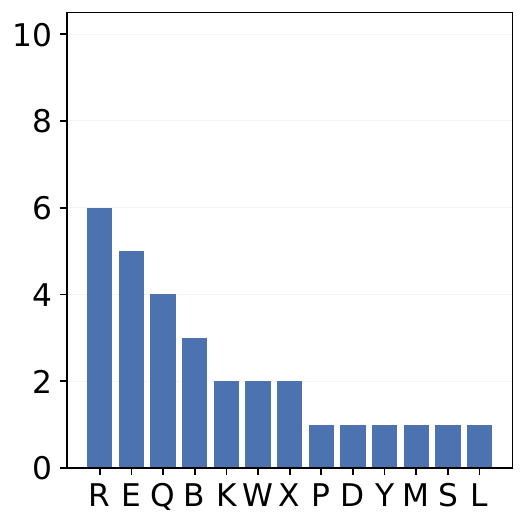}
        \vspace{-2mm}
        \caption{Reference: ``A'', ``G''}
    \end{subfigure}
    \begin{subfigure}{0.32\textwidth}
        \centering
        \includegraphics[width=0.9\linewidth]{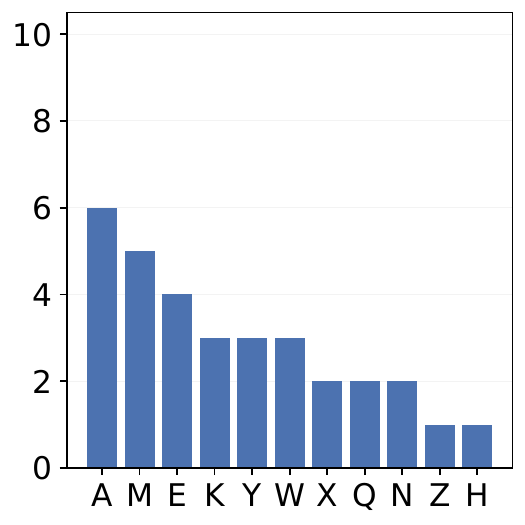}
        \vspace{-2mm}
        \caption{Reference: ``R'', ``S''}
    \end{subfigure}
    \vspace{-2mm}
    \caption{Histograms of the two additionally acquired reference characters aggregated over all test fonts for different two-reference settings ($K = 2$).}
    \label{fig:acquired_twoshot}
\end{figure}

\subsection{Trends in Acquired References}

We analyze overall trends in the characters acquired as additional references across the test fonts.
Fig.~\ref{fig:acquired_oneshot} shows histograms of the additionally acquired reference character aggregated over all test fonts for different single-reference settings ($K = 1$). Fig.~\ref{fig:acquired_oneshot}(a)–(c) correspond to the cases where the current reference is ``A'', ``B'', and ``C'', respectively.
When the current reference is ``A'', the proposed acquisition function frequently selects rounded characters such as ``R'', ``G'', and ``B'', as well as characters with distinctive terminals such as ``E'', ``T'', and ``F''.  
In contrast, when the current reference is ``B'', ``A'' is often selected, as it contains sharp corners and diagonal strokes.  
Notably, characters such as ``R'' and ``G'' that are frequently acquired when ``A'' is the current reference are rarely acquired when ``B'' is given. This is because their rounded parts are already covered by the current reference.

Fig.~\ref{fig:acquired_twoshot} presents histograms of the two additionally acquired reference characters aggregated over all test fonts for different two-reference settings ($K = 2$). Fig.~\ref{fig:acquired_twoshot}(a)–(c) correspond to the cases where the current references are (``A'', ``R''), (``A'', ``G''), and (``R'', ``S''), respectively.
Similarly to the single-reference setting, the proposed acquisition function tends to select characters that complement the parts missing from the current references.  
In Fig.~\ref{fig:acquired_twoshot}(a), where the references are ``A'' and ``R'', the rounded character ``Q'' is frequently selected.  
In Fig.~\ref{fig:acquired_twoshot}(b), where the references are ``A'' and ``G'', ``R'' is often acquired, as it contains a vertical stroke and a rightward curved component not sufficiently covered by the current references.  
In Fig.~\ref{fig:acquired_twoshot}(c), where the references are ``R'' and ``S'', angular characters such as ``A'' and ``M'' are frequently selected.

\begin{figure}[t]
    \centering
    \begin{subfigure}{0.32\textwidth}
        \centering
        \includegraphics[width=\linewidth]{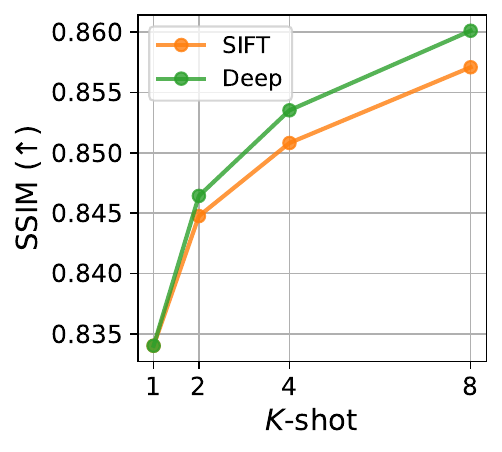}
        \vspace{-6mm}
        \caption{SSIM}
    \end{subfigure}
    \begin{subfigure}{0.32\textwidth}
        \centering
        \includegraphics[width=\linewidth]{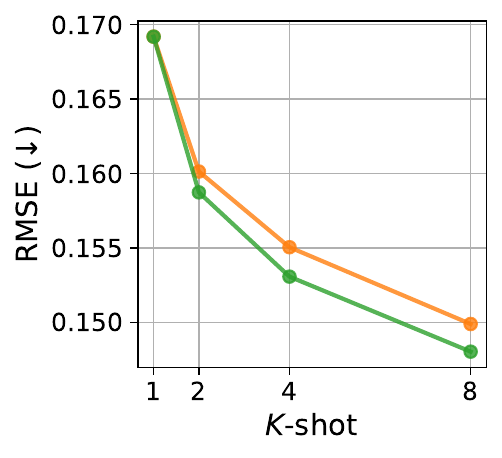}
        \vspace{-6mm}
        \caption{RMSE}
    \end{subfigure}
    \begin{subfigure}{0.32\textwidth}
        \centering
        \includegraphics[width=\linewidth]{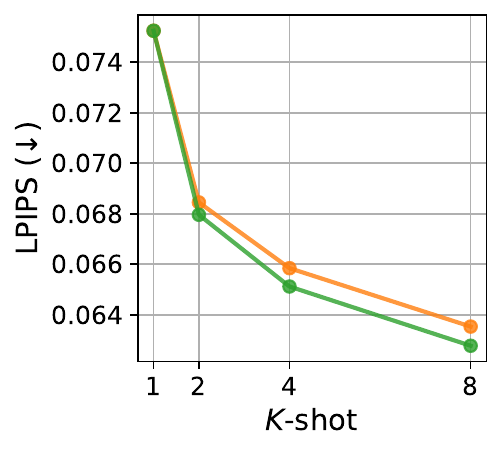}
        \vspace{-6mm}
        \caption{LPIPS}
    \end{subfigure}
        \vspace{-2mm}
    \caption{Performance comparison with different local part features.}
\label{fig:local_features}
\end{figure}

\begin{figure}[t]
    \centering
    \begin{subfigure}{0.32\textwidth}
        \centering
        \includegraphics[width=\linewidth]{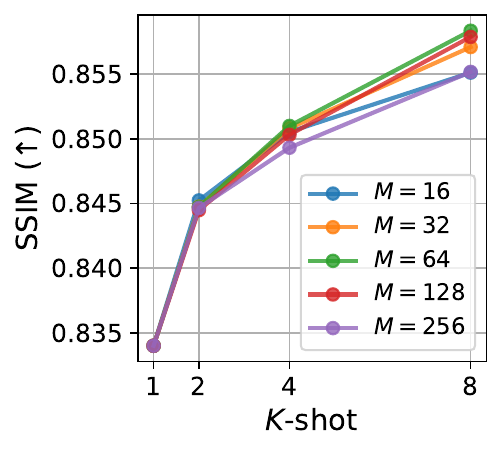}
        \vspace{-6mm}
        \caption{SSIM}
    \end{subfigure}
    \begin{subfigure}{0.32\textwidth}
        \centering
        \includegraphics[width=\linewidth]{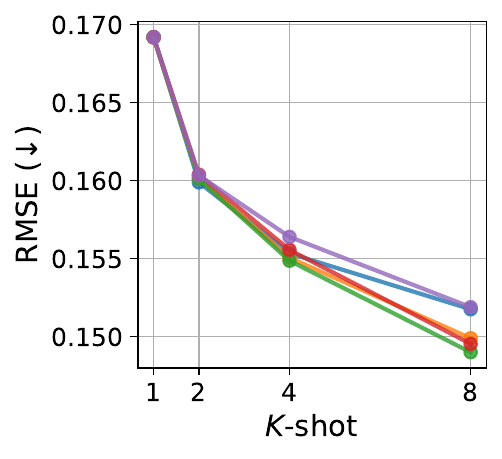}
        \vspace{-6mm}
        \caption{RMSE}
    \end{subfigure}
    \begin{subfigure}{0.32\textwidth}
        \centering
        \includegraphics[width=\linewidth]{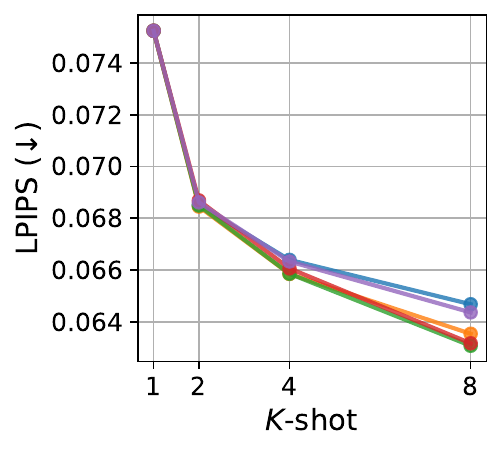}
        \vspace{-6mm}
        \caption{LPIPS}
    \end{subfigure}
        \vspace{-2mm}
    \caption{Performance comparison with different numbers of clusters $M$.}
\label{fig:num_clusters}
\end{figure}

\subsection{Effect of Local Part Features}
\label{subsec:local_features}

Since the part coverage used in our acquisition function is computed from local part descriptors, the choice of local features is expected to influence the effectiveness of the proposed method.
In this section, we investigate how different local part features affect active reference acquisition performance.

Specifically, we compare two choices of part descriptors: (i) SIFT descriptors and (ii) deep local descriptors extracted from a shallow layer of the style encoder, where each spatial location corresponds to a local region in the glyph image.
Fig.~\ref{fig:local_features} reports the results.

Across all metrics, using the deep local descriptors consistently achieves better generation quality than using SIFT.
While SIFT is a generic handcrafted descriptor, the style encoder is trained on the Google Fonts dataset and thus can capture font-specific local patterns more effectively.
These results suggest that our method benefits from stronger local representations: the more accurately the local part descriptors capture fine-grained glyph details, the more effective our part-coverage-based acquisition becomes.

\subsection{Effect of the Number of Clusters $M$}
\label{subsec:num_clusters}

Our acquisition function relies on a part histogram constructed by clustering local part descriptors; therefore, the choice of the number of clusters $M$ can affect the quality of the part-coverage representation.
In this section, we examine the impact of $M$ on acquisition performance.

We vary the number of clusters as $M\in\{16,32,64,128,256\}$ and report the results in Fig.~\ref{fig:num_clusters}.
We observe that $M=32$, $64$, and $128$ achieve comparable performance, while very small or very large values degrade performance across all metrics.
When $M$ is too small (e.g., $M=16$), distinct parts tend to be merged into the same cluster, which can obscure which parts are missing and may prevent the method from querying characters that contain truly uncovered parts.
Conversely, when $M$ is too large (e.g., $M=256$), similar parts can be split into multiple clusters, which may cause the acquisition to overestimate novelty and repeatedly query characters that largely overlap with already acquired parts.
These results suggest that $M$ should be set according to the complexity of local parts in the dataset to obtain a reliable part-coverage representation.

\section{Conclusion}
\label{sec:conclusion}

In this work, we introduced \emph{Active Reference Acquisition}, a new framework for few-shot font generation that enables actively querying a designer for additional reference glyphs.
To address this setting, we proposed a \emph{reference part-coverage-based acquisition function} that selects the next query character expected to maximize the part coverage of the current reference set.
By explicitly considering both the candidate character's part coverage and the parts already covered by the references, our acquisition strategy encourages querying characters that contain previously unseen parts, leading to more effective reference expansion.
Experiments on the Google Fonts dataset demonstrate the effectiveness of the proposed acquisition function.

As future work, we plan to validate the proposed framework with other few-shot font generation models and on larger-scale datasets covering diverse writing systems.
Since our acquisition function is largely model- and dataset-agnostic, we expect it to generalize to a wide range of generators and scripts, including non-Latin systems such as Chinese and Korean.
Another promising technical direction is to integrate generation uncertainty into the acquisition function when using probabilistic generative models such as diffusion models.
For example, by repeatedly generating the same candidate character and quantifying uncertainty in its local parts, such as whether the same part of the same character is generated as a straight stroke in some generations and as a curved stroke in others, the acquisition strategy could prioritize characters whose ambiguous or unstable parts are most likely to be resolved through additional designer input, leading to more informative reference acquisition.

\subsubsection*{Acknowledgements}
This work was supported by JST ACT-X Grant Number JPMJAX23CR.

%
\bibliographystyle{splncs04}
\bibliography{main}

\end{document}